# FastNet : An Efficient Architecture for Smart Devices

## John Olafenwa

*johnolafenwa@gmail.com*

## Moses Olafenwa

*guymodscientist@gmail.com*

## ABSTRACT

Inception [13] [14] [15] and the Resnet family of Convolutional Neural Network architectures [1] [2] [3] [5] have broken records in the past few years, but recent state of the art models have also incurred very high computational cost in terms of training, inference and model size. Making the deployment of these models on Edge devices, impractical. In light of this, we present a new novel architecture that is designed for high computational efficiency on both GPUs and CPUs, and is highly suited for deployment on Mobile Applications, Smart Cameras, Iot devices and controllers as well as low cost drones. Our architecture boasts competitive accuracies on standard Datasets even outperforming the original Resnet[1].

We present below the motivation for this research, the architecture of the network, single test accuracies on CIFAR 10 [20] and CIFAR 100 [20] , a detailed comparison with other well-known architectures and link to an implementation in Keras.

## MOTIVATION

Since Krizhevsky et al [4] broke records on imagenet in 2012, a tremendous amount of effort has been put into finding Computationally Efficient and highly accurate architectures. Tradeoff between computational efficiency and model accuracy has been a subject of great investigation. On the extreme side of very low model size, we have the Squeezenet architecture [9] which boasts AlexNet [4] level accuracy with 50 times less parameters and an ImageNet [21] model size of just 5 mb, on the extreme side of Accuracy, we have the Wide Resnet [5] which boasts the state of the art accuracy on most of the standard datasets, but with an ImageNet [21] model size of 260 mb, making them usable only for cloud services.

Smartphones have become a core part our lives and in few years, Internet of Things devices would become a core part of our homes and industries. Going forward, all our mobile applications and IoT devices would need to have Intelligence infused into them. Edge devices relying on cloud hosted models would not suffice for the modern AI needs, network latency, failure and bandwidth costs makes intelligent applications relying solely on cloud services, not fully reliable. A number of techniques have been employed to address these, Network pruning, reduction in floating point precision and the Squeezenet Architecture [9], however, the accuracies of this models are not often comparable to the models deployed on the cloud.

To build truly reliable intelligent edge devices, there is need for architectures that are very accurate but can be fit in edge devices.

# FASTNET

We are presenting a new architecture, named "FastNet"

FastNet is a 15 layer Convolutional Network that explores the concepts of medium depth and medium network width.

It boasts accuracies of 93.98% on CIFAR10 and 70.81% on CIFAR100, with only 1.6M parameters, coupled with high training and inference speed. Using Keras with Tensorflow [24] backend, training on CIFAR100 takes approximately 3 hours on a single NVIDIA P100.

Recent models have explored absolute depth, while Wide Resnet [15] uses high depth with great width. However, the results of their works demonstrates that the rate of increase of depth, while directly proportional to increase in accuracy, is marginally disproportionate. Increasing depth greatly often slows down training and inference with little gain in accuracy. This questions the benefits of very high depth in neural networks. And challenges us to rather seek more efficient means of improving model accuracy.

FASTNET IS STRUCTURED AS FOLLOWS:

# UnitCell

Each layer is made up of BatchNormalization followed by RELU Activation and finally Convolution. Batch Normalization introduced [11] is used to normalize all feature maps to have zero mean and unit variance. This helps to correct Internal Covariate Shift, a phenomenon that results from shift in the distribution of the activation maps as a result of changing parameters. Batch Normalization is expressed formally as

$$y_i \leftarrow \gamma \frac{x_i - \mu}{\sqrt{\sigma^2 - \epsilon}} + \beta$$

Note that $\gamma$ and $\beta$ are learnable parameters

They are estimated during training via stochastic gradient descent.

Rectified Linear units are used as the activation function, as they are very efficient to compute and has been empirically proved to be highly effective.

They take the form

$$y = \max(0, x)$$

It effectively threshold activations at 0.

### 3 x 3 Convolutions

The first 12 layers of FastNet are constructed with 3 x 3 convolutions. This is in line with recent studies that has shown the effectiveness of 3 x 3 convolutions, first they capture a sufficiently wide region to be able to detect abstract patterns properly, and they are computionally efficient .

Christian Szegedy et al [14] demonstrated in "Rethinking The Inception Architecture" that two 3 x 3 convolutions stacked upon each other performs similarly to a single 5 x 5 convolution while being 28% more efficient. Hence, specially in this case where we are attempting to optimize for computational efficiency, 3 x 3 convolution is the reasonable choice.

### Final 1 X 1 Convolutions:

We avoid using fully connected layers in the final layers of the network as they add a very large number of parameters to the network. Concurrent with recent practices, we replace fully connected layers with a stack of 3, 1 x 1 Convolutions followed by a GlobalAveragePooling layer. The output of the AveragePooling layer is fed directly into the softmax layer.

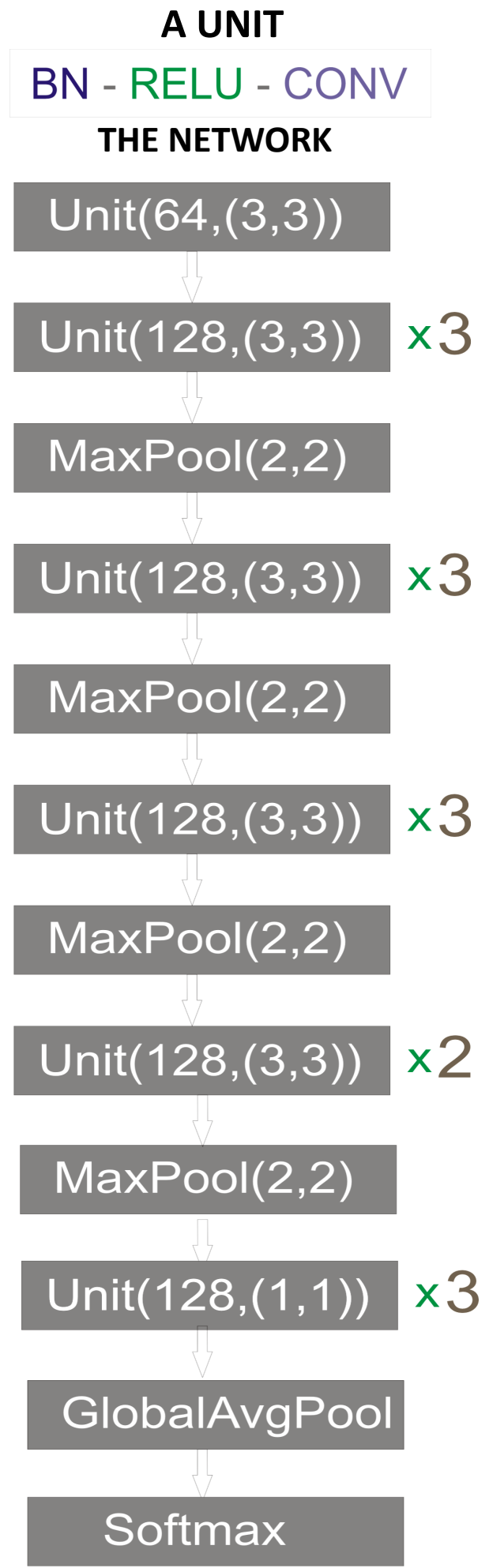

## Late Downsampling:

Pooliing feature maps have been demonstrated to enable Covnets generalize properly, making Covnets highly invariant to the presentation of the image. However, pooling at early layers can lead to loss of valuable information about the structure of the image, hence, we perform the first pooling after the first four Convolutional layers. We use MaxPooling with a pool size of 2.

# Summary

FastNet begins with a 64 channels 3 x 3 Conv Unit layer, followed by three 128 channel layers with similar composition as the first, Down sampling follows, then 3 Unit Layers follows, and we down sample again, another three layers followed by down sampling is used again, after which is 2 layers followed by down sampling. Finally, we stack two 1 x 1 Unit Cells followed by Global Average Pooling.

This is fed into the softmax layer.

## AVOIDING PARALLEL LAYERS

Inception [15] and FractalNet [6], both which performs excellently without using Residual Connections, make great use of layer parallelization techniques. This is very effective on GPUs, because they are really good at multi-threading, however, on Edge Devices, inference is primarily by CPUs. It is well known that CPUs have limited multi-threading capabilities, hence we explicitly avoid using this technique.

## SIMPLICITY OF DESIGN

A model is only useful enough when many developers can make use of it. Recent architectures have deviated from the simplicity of VGG Net [16], hence, a lot of ML engineers with limited knowledge find it hard to replicate these architectures. FastNet follows the VGG style and is very simple to implement in any Deep Learning framework.

## RELATED WORKS

Kaiming et al [1] won ImageNet 2015 with their ground breaking work on Residual Connections. They further improved on their work in 2016 with "Identity Mappings in Deep Residual Networks" [2] Almost all architectures since then have been based upon these framework, in fact, most new architectures are adaptations of the original Residual Framework, with the exception of Fractal Net [6] . Stochastic Depth [3], Wide Resnets [5] and Share Resnet [23] have all sought to improve the speed of Residual Networks. However, all of these still focus on exploration of depth, while they represent significant improvements over the original Resnet [1], they are still unsuitable for deployment on Edge devices.

Fractal Net [6] is a great deviation from the Resnet Family. It is a highly impressive work that proved that Residual Connections are not an absolute requirement for improving accuracy. It essentially uses highly parallel layers, this is very central to the design of the architecture. This is an excellent fit for GPUs, but such layer parallelization is not good for CPU dependent Edge Devices. Limited cores and thread context switching would fundamentally hamper performance of these network on low end devices.

FastNet is far more efficient that the art of the art architectures while being close in terms of accuracy. It even outperforms a number of well-known architectures including the original Resnet [1].

## COMPARISONS

FastNet outperforms a number of popular network architectures, it also outperforms 110 layer Resnet [1] on CIFAR 10. Even though, a few state of the art architectures boasts better accuracies, but that comes at very high computational cost that renders them unusable on Edge devices. we shall now make clear comparisons with some existing architectures. Note, these comparisons are simply for proofs, we have the utmost respect for the authors of these works, they are pioneers from which we have greatly learned.

| ARCHITECTURE | CIFAR 10 | CIFAR 100 | Params |
| --- | --- | --- | --- |
| **FastNet (Ours)** | **93.98** | **70.81** | **1.6 M** |
| Network In Network [12] | 91.19 | 64.32 | 1 M |
| ALL CNN [17] | 92.75 | 66.29 | 1.3 M |
| MaxOut [8] | 90.62 | 65.46 | 6 M |
| Resnet 11O [1] | 93.57 | 74.84 | 1.7 M |
| Wide Resnet [5] | 95.83 | 79.5 | 36 M |
| VGG Net [16] | 91.4 | - | 138 M |
| Fractal Net [6] | 95.4 | 76.27 | 38.6 M |
| Stochastic Depth [3] | 94.77 | 75.42 | 1.7 M |
| Fractional Max Pooling [19] | - | 68.55 | - |
| Fractional Max Pooling With Large Aug. (12 tests) | 95.5 | 73.61 | 50 M |

**Note:** It can be observed that the number of parameters for Resnet 110 and Stochastic Depth are just slightly above FastNet, however, the actual difference in Model size and performance speed is much higher by a significant margin beyond what the difference in the number of parameters tells. These networks use very thin layers that makes their parameters less but use very high depth that significantly slows them down.

# EXPERIMENT SETUP

Experiments on CIFAR 10 and CIFAR 100 were conducted using data augmentation techniques similar to Wide Resnet [5]. Adam optimizer with an initial learning rate of 0.001 was used to train the network and the learning rate was divided at 80,120, 160 and 180 epochs. Weights were initialized with he_normal as proposed by Kaiming et al [7]. All experiments ran for a total of 200 epochs. Code was written in keras and can be found on this Github repo. (https://github.com/johnolafenwa/FastNet)

We used the standard softmax cross entropy loss as our loss function.

It takes the form

$$L = -\log\left(\frac{e^{x_j}}{\sum e^{x_i}}\right)$$

Where $j$ is the index of the correct class.

# CONCLUSIONS

In view of the results of our experiments, we draw the following conclusions.

1. Our architecture, FastNet is highly suited for Edge devices, and is highly optimized for all CPU dependent devices.
2. Simpler architectures when properly designed can outperform complex architectures.
3. Medium depth with medium width networks can perform well with much lesser computational cost.

We also believe, based on these results, that Ultra Deep Networks are not an absolute necessity for building an efficient architecture. The future of CNNs is not going to be determined by arbitrary increase in depth but rather a conscious effort to optimize the hyper-parameters of an architecture as well as new better dimensionality reduction techniques, improved activation functions and maybe someday, an effective replacement for the convolution layer itself. A notable example is the introduction of Batch Normalization by Ioffe et al [11]. Batch Normalization greatly improved both the accuracy and efficiency of existing neural networks, unlike deeper layers that slightly improves accuracy but greatly reduces computational efficiency. Such great ideas need to be pursued with great vigor.

We hope with further resources to conduct our research, we shall be able to further improve both the accuracy and efficiency of FastNet.

## IMAGENET

As two independent researchers, we have limited resources and cannot at present conduct experiments on ImageNet. However, good results on CIFAR 10 and CIFAR 100 always yields good results on ImageNet as well, this makes us highly optimistic about FastNet. We hope with availability of more resources in the near future, we shall be able to significantly improve upon this baseline work.

## FURTHER WORK

The performance of FastNet on Imagenet still needs to be evaluated to ensure fair comparison to other Architectures. Also, we strongly believe, with further research, the computational efficiency and accuracy of this model can be increased.

To make computer Vision available to everyone, there is absolute need for great research not just into more accurate models setting new state of the art accuracy, but on highly efficient models that can work well on low cost edge devices. If we search deeply in the direction of efficiency, we can someday build computer vision systems with near human efficiency.

Artificial General Intelligence would only be fully realized when we are able to build intelligent systems that are both accurate and efficient.

## ABOUT THE AUTHORS

### JOHN OLAFENWA

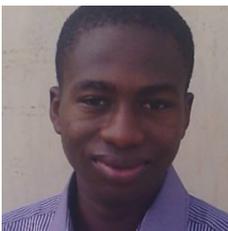

A self-taught computer programmer, Neural Networks Blogger and Computer vision researcher. Skilled in Building Android applications and Native software. Can develop software with Java, Python and C#. Very passionate about transforming lives through highly efficient neural networks. Studies reInforcement learning at leisure time.

Email: johnolafenwa@gmail.com,

Website: john.specpal.science

Twitter: @johnolafenwa

### MOSES OLAFENWA

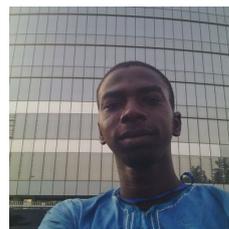

A self-Taught computer programmer, Cloud and Internet Logistics expert. Skilled in developing Android applications Web portals and Desktop software. Can code in Java, Python and PHP. A Deep Neural Network practitioner with a vision to make the world better via Artificial Intelligence. A lover of Big Data.

Email: guymodscientist@gmail.com,

Website: moses.specpal.science

Twitter: @OlafenwaMoses